\documentclass[11pt,a4paper]{article}
\usepackage[hyperref]{ranlp2025}
\usepackage{times}
\usepackage{latexsym}

\usepackage{microtype}
\usepackage{pifont}
\usepackage{xspace}
\usepackage{float} 
\usepackage{pifont} 
\usepackage{bbold}
\usepackage{amssymb}
\usepackage{booktabs}
\usepackage{multirow}
\usepackage{graphicx}
\usepackage{etoolbox}
\usepackage{float}

\aclfinalcopy 

\title{Automatic Fact-checking in English and Telugu}

\author{
\begin{tabular}{ccc}
  Ravi Kiran Chikkala\textsuperscript{1,2} &
  Tatiana Anikina\textsuperscript{3} &
  Natalia Skachkova\textsuperscript{3} \\
  Ivan Vykopal\textsuperscript{4,5} &
  Rodrigo Agerri\textsuperscript{2} &
  Josef van Genabith\textsuperscript{3} \\
\end{tabular} \\
\textsuperscript{1} Saarland University \\
\textsuperscript{2} University of the Basque Country \\
\textsuperscript{3} German Research Center for Artificial Intelligence, Saarland Informatics Campus \\
\textsuperscript{4} Faculty of Information Technology, Brno University of Technology, Brno, Czech Republic \\
\textsuperscript{5} Kempelen Institute of Intelligent Technologies, Bratislava, Slovakia \\
\texttt{rach00004@teams.uni-saarland.de} \\
\texttt{\{tatiana.anikina,natalia.skachkova,josef.van\_genabith\}@dfki.de} \\
\texttt{ivan.vykopal@kinit.sk} \\
\texttt{rodrigo.agerri@ehu.eus} \\
}

\date{}

\begin{document}
\maketitle
\begin{abstract}
False information poses a significant global challenge, and manually verifying claims is a time-consuming and resource-intensive process. In this research paper, we experiment with different approaches to investigate the effectiveness of large language models (LLMs) in classifying factual claims by their veracity and generating justifications in English and Telugu. The key contributions of this work include the creation of a bilingual English-Telugu dataset and the benchmarking of different veracity classification approaches based on LLMs.

\end{abstract}

\section{Introduction}

In today's technological world, claim verification plays an important role \cite{zhang-gao-2023-towards}, which aims to assess the veracity of claims as ``true'' or ``false'' by validating them against trustworthy sources \cite{Panchendrarajan_2024}. This is necessary to combat false information, especially in multilingual countries such as India, where false information can be propagated in multiple languages via translation technology \cite{Quelle2025}. According to \citet{pradeep-etal-2021-scientific}, claim verification involves three key steps: (1) retrieval of documents, (2) rationale selection, and (3) label prediction. Currently, multilingual LLMs significantly improve the claim verification process \cite{10.5555/3666122.3668964} compared to traditional approaches such as manual fact-checking and simple machine learning classifiers. These language models not only evaluate claims, but also provide justifications, thereby offering a level of explanation that traditional natural language processing (NLP) approaches often lack \cite{Dmonte2024ClaimVI}. To date, most of the work on claim verification in fact-checking has been performed in English. In this work, we address this shortcoming by creating a new fact-checking dataset in Telugu, allowing for large-scale experimentation in Telugu, a language spoken by over 200 million people in the world \cite{Mallareddy2012EvolutionOT}. We achieve this by translating our manually created English dataset into Telugu, resulting in a bilingual English–Telugu dataset that supports multilingual claim verification. Furthermore, LLMs pose several limitations, such as tendencies to hallucinate \cite{li-etal-2024-dawn}, they exhibit biases \cite{lin-etal-2025-investigating}, smaller models may operate within limited context windows \cite{ratner-etal-2023-parallel}, and models may rely on knowledge that may be outdated due to cutoff dates \cite{cheng2024dated}. In order to address these challenges, we use Retrieval-Augmented Generation (RAG)  \cite{NEURIPS2020_6b493230} with different components, such as prompt compression \cite{li-etal-2025-prompt}, document re-ranking \cite{hui-etal-2022-ed2lm} and query rewriting \cite{ma-etal-2023-query}. 

We explore two research questions.

    \begin{itemize}
        \item[] \textit{\textbf{RQ1:}} How well do LLMs classify domain-specific claims in English versus Telugu?
        
        \item[] \textit{\textbf{RQ2:}} How do different models and approaches impact the quality of justifications provided by LLMs in a English-Telugu multilingual setting?
    \end{itemize}


To address these research questions, we introduce a new dataset named \textbf{Preethi}\footnote{We make our complete dataset available at \url{https://huggingface.co/datasets/Blue7Bird/Preethi_dataset}} that covers both English and Telugu. Our experiments demonstrate that RAG-based approach, achieves the highest claim verification scores in both English and Telugu. For justification generation, RAG-based approach obtains the best average score for English, while Simple Prompting achieves the highest average score for Telugu.

\section{Related Work}


\subsection{Datasets Related to Indian Languages}

Several datasets have been proposed for detecting false information in the Indian context. \citet{article} introduce the Indian Fake News Dataset (IFND), a monolingual English dataset comprising 56,714 claims across various categories relevant to the Indian context. Each claim in IFND is labeled as ``true'' or ``fake''. Similarly, \citet{gupta-srikumar-2021-x} develop the X-Fact dataset, which includes 31,189 claims and supports multiple Indian languages-though not Telugu. X-Fact has five labels ``true", ``mostly-true", ``partly-true", ``mostly-
false", and ``false". \citet{Singhal_Shah_Kumaraguru_2022} annotate the Fact Drill dataset, which comprises 22,435 false claims in 13 Indian regional languages, including fewer than 2,000 samples in Telugu. However, the dataset is not publicly available. \citet{mittal-etal-2023-lost} present the X-CLAIM dataset, which focuses on the identification of claims in multilingual social media posts. X-CLAIM contains 7,000 real-world claims across five Indian regional languages and English, but only 107 Telugu samples in its test set. \citet{10.5555/3666122.3668964} develop the AVeriTeC dataset, comprising 4,568 real-world claims in English. Each claim in AVeriTeC is classified into one of the four labels: ``supported", ``refuted", ``not enough evidence" and ``conflicting evidence/cherry-picking".   \citet{Raja2023FakeND} create the Dravidian Fake News Dataset (DFND), which consists of 26,000 news articles in Telugu, Tamil, Kannada, and Malayalam, annotated with binary labels: ``true" or ``fake". However, the DFND is not open source, which poses challenges for reproducibility and further research.

Although some of these datasets support claim verification to varying degrees in Indian languages, none, except AVeriTeC, include human-annotated justifications and Question Answer (QA) pairs. Yet AVeriTeC is not designed for the Indian context. This highlights a research gap: the absence of open source, human-annotated QA pairs, and justification-rich resources for misinformation detection in low-resource Indian languages such as Telugu for the Indian context.

\subsection{RAG and Other Approaches with LLMs}

Recent advances in claim verification have used LLMs and RAG frameworks for claim verification processes \cite{Dmonte2024ClaimVI}. \citet{singal-etal-2024-evidence} develop a RAG pipeline that extracts relevant evidence sentences from a knowledge base, which are then passed into an LLM for classification. \citet{yue-etal-2024-retrieval} introduce a Retrieval‑Augmented Fact Verification framework through the synthesis of contrasting arguments (RAFTS) to determine the veracity of the claim. \citet{Katranidis2024FaaFFA} propose Facts as a Function approach (FaaF), which is based on RAG, to evaluate the factual accuracy of the text generated by LLMs.
\citet{Vykopal2024GenerativeLL} present a comprehensive review of claim verification frameworks that use LLMs, focusing on methods such as RAG and fine-tuning. Our work is different from previous research, as we implement a RAG pipeline that enhances LLMs' fact-checking capability, using Automatic Scraping, integrating both foundational and Advanced RAG components. We use Really Simple Syndication (RSS)~\cite{wikipediaRSS} feeds from reputable Indian news sources, chosen for their longstanding credibility and wide readership, to access up-to-date information to assess new claims, as LLMs' have knowledge cutoff dates and may contain outdated information.

\section{Preethi Dataset}

\begin{figure}[h] \centering \includegraphics[width=0.5\textwidth]{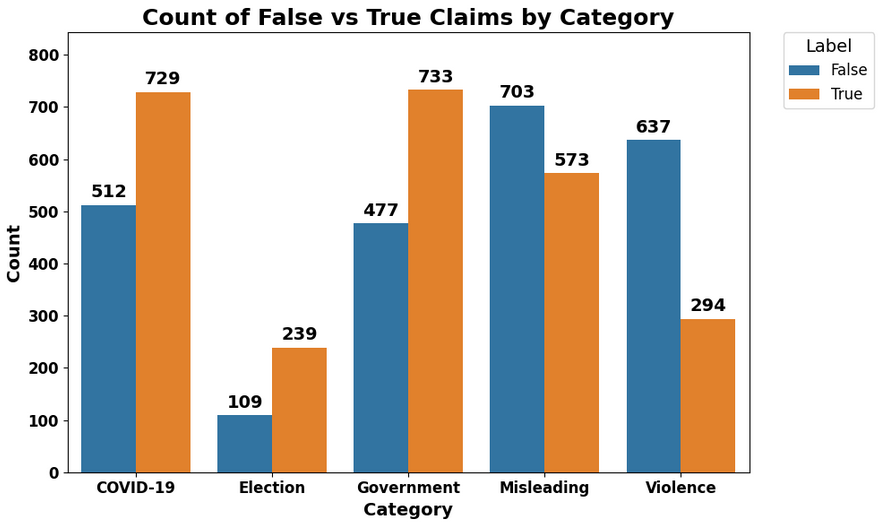} \caption{Statistical information of the Preethi dataset about true and false claims across five categories} \label{fig:preethi_dataset} \end{figure}

In this research work, we have created the Preethi dataset, which is based on the publicly available IFND \cite{ifndDataset}. A claim, as defined by \citet{Panchendrarajan_2024}, is a statement that can be verified against evidence. The IFND has several inconsistencies such as incomplete claims, non-claims, questions, and entries with multiple claims; these inconsistencies compromise its overall quality as these claims cannot be verified against evidence, see Table \ref{tab:inconsistencies_in_IFND} for examples of inconsistent claims. We chose IFND because it is publicly available and its inconsistencies highlight the need for a refined and higher-quality resource, an opportunity we address through the creation of Preethi dataset. We have manually annotated a dataset of 5,006 claims in English with five topics from IFND, namely \textit{Covid-19}, \textit{Election}, \textit{Government}, \textit{Misleading}, and \textit{Violence}. Statistical details of the Preethi dataset are presented in Figure \ref{fig:preethi_dataset}. The Preethi dataset is not a strict subset of IFND. Of the 2,568 true claims, 2,500 are sourced from IFND. Among the 2,438 false claims, 2,435 are collected from fact-checking websites. Following, \citet{10.1080/23808985.2019.1602782} we treat partially true claims from fact-checking sources as false, given their potential to spread misinformation similar to fully false claims. To reconstruct complete claims from inconsistent IFND entries, we use their original sources, identified via Google Web Search \cite{google} and, when necessary, Microsoft Copilot \cite{copilot}.

\begin{table}[ht]
\scriptsize
\centering
\begin{tabular}{|p{5.3cm}|c|}
\hline
\textbf{Claim} & \textbf{Inconsistency} \\
\hline
This Video Is Not Of UP Police Chasing A......... & Incomplete\\
\hline
Did Israel bomb Iranian nuclear facilities? & Question \\
\hline
Drop, Don't Extend It & Non-claim\\
\hline
\parbox{4.6cm}{India's Ministry of Culture has NOT announced a relief....Fact Check: Chill. Iceland hasn't declared religions as weapons of mass destruction} & Multiple claims\\
\hline
\end{tabular}
\caption{Inconsistent claims  in IFND}
\label{tab:inconsistencies_in_IFND}
\end{table}

Inspired by the AVeriTeC, we provide additional metadata for each claim, including supporting documents from the Web, the date of the claim, gold justifications, and gold QA pairs. 
Gold justifications and gold QA pairs are created manually based on the information in the supporting documents. To maintain the quality of the dataset, we have involved three annotators who were trained via detailed guidelines. We achieve a Cohen’s Kappa agreement score of 80\% for claim veracity labels and 75\% for boolean QA pairs, indicating substantial inter-annotator agreement. In addition, all abstractive and extractive QA pairs are manually checked by annotators for correctness and relevance by verifying them against supporting documents. To make our dataset available in Telugu, we translate the English dataset using the Google Translate API \cite{googletranslate}. To assess the quality of the translated data, we perform a back-translation from Telugu to English and compare it with the original English version. This results in a BLEU \cite{papineni-etal-2002-bleu} score of 0.255 and a METEOR \cite{banerjee-lavie-2005-meteor} score of 0.659, indicating moderate consistency between the original and back-translated texts. However, the raw machine translations are not directly used in our experiments. Instead, three native Telugu speakers have manually post-edited the machine translated output and removed the syntactic and semantic errors. The final Telugu dataset is used for experiments, ensuring high-quality translations and minimizing the potential bias introduced by machine translation errors. We calculate post-edits by comparing the initial machine-translated Telugu dataset with the final manually annotated Telugu dataset using Pyter \cite{pyter} to measure the translation error rate (TER) \cite{snover-etal-2006-study}. A total of 31,465 post-edits are made. 
Table~\ref{tab:dataset_comparison_small} compares Preethi dataset to the existing benchmark datasets.

\begin{table}[!ht]
\scriptsize
\centering
\begin{tabular}{|l|c|c|c|}
\hline
\textbf{Dataset} & \textbf{Justifications} & \textbf{Supports Telugu} & \textbf{QA Pairs} \\
\hline
X-CLAIM       & \ding{55}  & $\checkmark$   & \ding{55} \\
AVeriTeC      & $\checkmark$ & \ding{55}  & $\checkmark$ \\
DFND          & \ding{55}  & $\checkmark$  & \ding{55} \\
IFND          & \ding{55}  & \ding{55}  & \ding{55} \\
X-Fact        & \ding{55}  & \ding{55}  & \ding{55} \\
Fact Drill    & \ding{55}  & $\checkmark$  &  \ding{55} \\
\textbf{Preethi (ours)} & \textbf{$\checkmark$} & \textbf{$\checkmark$} & \textbf{$\checkmark$} \\
\hline
\end{tabular}
\caption{Comparison of Preethi dataset with benchmark datasets.}
\label{tab:dataset_comparison_small}
\end{table}

\subsection{QA Pairs} 
\label{sec:qa-pairs}
Each claim in our dataset has three manually created QA pair types; see Table \ref{tab:eiffel_qa_comparison} for examples.

\textbf{Boolean:} Our dataset contains 4,010 indirect and 996 direct boolean QA pairs. Direct QA pairs rephrase the claim itself as a yes/no question, while indirect QA pairs pose a related yes/no question that helps verify the validity of the claim.



\textbf{Abstractive:} QA pairs are created by summariz-
ing the relevant information about the claim.

\textbf{Extractive:} QA pairs, in which the answer is a
direct snippet or a phrase taken word-for-word.

\begin{table}[ht]
\centering
\scriptsize 
\renewcommand{\arraystretch}{1.4}
\resizebox{\linewidth}{!}{%
\begin{tabular}{|l|l|}
\hline
\textbf{Claim} & \textit{The Eiffel Tower is in London} \\
\hline
\textbf{QA Type}       & \textbf{Question(Q) \& Answer(A)} \\
\hline
Direct Boolean         & Q: Is the Eiffel Tower in London? A: No \\
Indirect Boolean       & Q: Is the Eiffel Tower in France? A: Yes \\
\hline
Abstractive            & Q: What is the Eiffel Tower? A: a well known monument.... \\
\hline
Extractive             & Q: Where is the Eiffel Tower? A: Paris, France. \\
\hline
\end{tabular}
} 
\caption{Boolean, Abstractive and Extractive QA pairs}
\label{tab:eiffel_qa_comparison}
\end{table}

\section{Methodology and Experiments}
\begin{figure}[h!] \centering \includegraphics[width=0.4\textwidth]{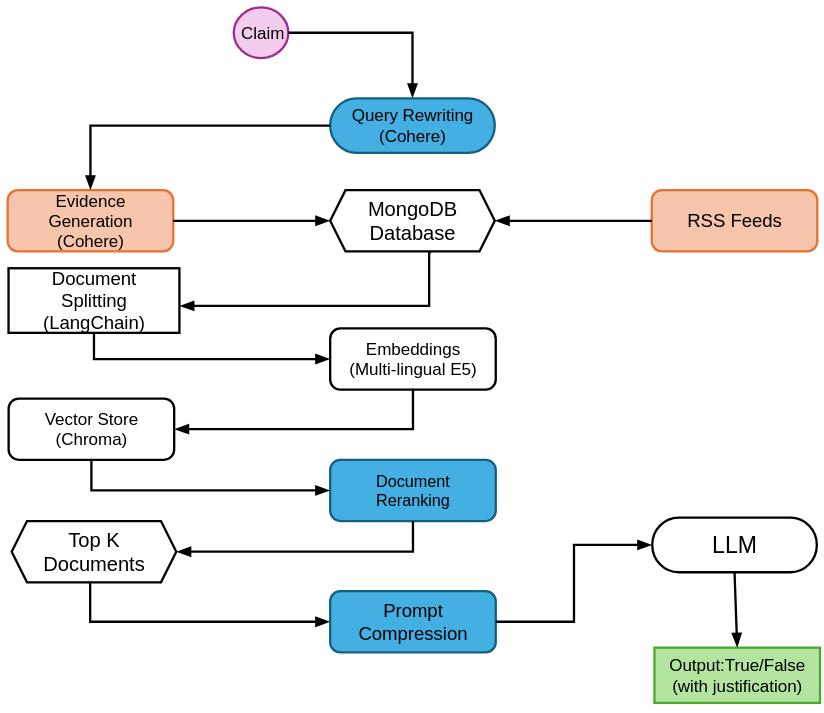} \caption{RAG Approaches} \label{fig:RAG} \end{figure}

This section discusses different approaches that are used in our experiments: \textit{1) Simple Prompting} and \textit{RAG} approaches that include \textit{2) Naive RAG};
 \textit{3) Advanced RAG} and \textit{4) Automatic Scraping}.  In our experiments, we use gold justifications to evaluate the justifications generated by LLMs and gold QA pairs to assess the quality of QA pairs generated by LLMs. For claim veracity evaluation, we use F1 score. In order to evaluate the justifications generated by LLMs, we use METEOR, ROUGE-L (R-L) \cite{lin-2004-rouge}, ChrF \cite{popovic-2015-chrf}, BERTScore \cite{bert-score} and BLEURT \cite{sellam-etal-2020-bleurt}. We make our complete code  \footnote{\url{https://github.com/formallinguist/Automatic-Fact-Checking}} and additional details public.

\subsection{Simple Prompting}

In Simple Prompting, we use a zero-shot approach \cite{DBLP:conf/iclr/WeiBZGYLDDL22}, where the LLM relies solely on its pre-trained knowledge and general language understanding to classify only claims, operating without any additional supporting documents. In this approach the LLM is given a claim as input, and it is tasked to classify a claim as ``false" or ``true" and provide reasoning or justification for its decision. Without such explanations, the classification may appear arbitrary or unsupported. For the experiments, we consider the Simple Prompting approach as a baseline. 


\subsection{RAG Approaches}

Since LLMs are not updated regularly and have a fixed knowledge cut-off date, they may hallucinate. To address this, we use RAG. In order to find supporting documents for claims, we use the Cohere c4ai-command-r7b-12-2024 \cite{cohere} model for English and Telugu. To handle the large number of new claims that appear every day, we use RSS feeds. These feeds are updated regularly by different on-line news sources, providing up-to-date information. To manage this data, we choose the MongoDB \cite{mongodb} database for our experiments. It is a NoSQL database suitable for unstructured data, making it ideal for storing RSS feeds and supporting documents retrieved by Cohere. We collect RSS feeds from reliable Indian news sources such as NDTV \cite{ndtv} for English and Eenadu \cite{eenadu} for Telugu. Chroma \cite{chromadb}, a vector database, is used to store documents using vector representations.

\subsubsection{Naive RAG}

In the Naive RAG approach, as shown in Figure \ref{fig:RAG} (excluding the steps highlighted in blue), the process unfolds as follows: 

\textit{Step 1:} Cohere c4ai-command-r7b-12-2024 model is prompted to provide supporting documents for a given claim. These documents are then stored in MongoDB. 


\textit{Step 2:} The MongoDB retriever, which uses string matching, identifies, and retrieves documents relevant to the claim. The retrieved documents are processed through the LangChain text splitter \cite{langchain_TextSplitter}, which divides documents into smaller segments.

\textit{Step 3:} These segments are converted to vector embeddings using multilingual E5 Text embeddings \cite{Wang2024MultilingualET}. These embeddings are then stored in Chroma.

\textit{Step 5:} Cosine similarity is used to compare the embeddings of the claim with the documents stored in Chroma. The top three documents with the highest cosine similarity scores are retrieved from Chroma and used as evidence for the LLM.

\textit{Step 6:} Finally, the LLM analyzes the evidence in the context of the claim and classifies the claim as true or false, along with justifications for its decision.


\subsubsection{Advanced RAG}

Advanced RAG is similar to Naive RAG but with additional components such as query re-writing, document re-ranking, and prompt compression. In Figure \ref{fig:RAG} the additional components of Advanced RAG are highlighted in blue.

\textbf{Query Re-writing:} For query re-writing, we use the Cohere c4ai-command-r7b-12-2024 model, which modifies the original claim to improve its quality for better retrieval of documents. This includes correcting spelling errors, rephrasing, or adding additional context to a claim for better understanding. See Table \ref{tab:rewritten_claims} for examples. According to \citet{skitalinskaya-wachsmuth-2023-revise}, the criteria for re-writing a claim include maintaining syntactic and semantic coherence, being grammatically correct, and removing ambiguity. A good claim is precise, includes relevant context, and is not ambiguous. In our experiments, we observe that 50\% - 60\% of claims undergo this process. We calculate this using string matching. We have manually verified 50 claims in English and Telugu to check the quality of the re-written claims. We observe that re-written claims in English are syntactically and semantically coherent, while Telugu re-written claims have grammatical errors. 


\begin{table*}[!ht]
\small
\centering
\begin{tabular}{|p{0.5cm}|p{5.5cm}|p{6cm}|}
\hline
\textbf{No.} & \textbf{Original Claim} & \textbf{Re-written Claim} \\
\hline
1 & Jharkhand new hotspot of illicit opium cultivation: NCB & The NCB reports significant opium cultivation in Jharkhand, identifying it as a potential hotspot. \\
\hline

2 & Govt confident of privatising Air India, BPCL by first half of 2021-22 divestment secretary & The Indian Government's Divestment Strategy: Privatization of Air India and BPCL by 2022 and the Secretary's Statement on Future Plans. \\
\hline
\end{tabular}
\caption{Comparison of original and re-written claims}
\label{tab:rewritten_claims}
\end{table*}

\textbf{Document Re-ranking:}  For document re-ranking, we use bert-multilingual-passage-reranking-msmarco \cite{amberoad-bert-multilingual-passage-reranking-msmarco}. It calculates the relevance of each document with respect to the claim and then sorts the documents by the scores to determine the best matches. This ensures the documents that are most relevant for the claim are ranked higher for further processing. Unlike cosine similarity in the Naive RAG approach, which only compares vector proximity, here, the CrossEncoder evaluates the relationship between the claim and the document in context. The top three re-ranked documents are considered for further processing.

\textbf{Prompt Compression:} For prompt compression, we use the Cohere  c4ai-command-r7b-12-2024 model. This involves reducing the length of a prompt while retaining its most important information. This helps in scenarios where there is a limited context window for an LLM. 


\subsubsection{Automatic Scraping}

In Automatic Scraping, we extract content from URL in the supporting documents of the Preethi dataset using BeautifulSoup (BS4) \cite{beautifulsoup4}. To overcome the limitation of the context window of the LLMs, we use a sentence-transformers/paraphrase-multilingual-mpnet-base-v2 \cite{reimers-gurevych-2019-sentence}. This model identifies the most relevant sentences from the supporting documents by comparing their semantic similarity to the given claim. We retrieve up to 3,000 characters of content that are most relevant to the claim. This selected content is then used as the context for the LLM and is referred to as the \textit{refined context}. This approach is repeatable with new data if the claim and its URL are available.


\subsection{Evaluation of QA pairs}

In claim verification, asking good questions is crucial \cite{10.5555/3666122.3668964}. To assess the quality of QA pairs generated by LLMs, we calculate their similarity to gold-standard QA pairs. We use an in-context learning approach \cite{dong-etal-2024-survey}, where the gold QA pair serve as reference to guide the LLM in generating boolean, abstractive, and extractive QA pairs in the desired format for a claim. For evaluation, we follow the approach of \citet{10.5555/3666122.3668964}, we first compute METEOR scores and then apply the Hungarian algorithm \cite{kuhn1955hungarian} to identify the optimal one-to-one matching between LLM-generated and gold-standard QA pairs by maximizing METEOR scores. Table \ref{tab:qa_pairs} provides English and Telugu scores. 


\subsection{Experiments}

We use the models listed in Table~\ref{tab:models} for experiments.

\begin{table}[H]
\centering
\footnotesize
\begin{tabular}{@{}lc l@{}}  
\toprule
\textbf{Versions of models} & \textbf{Parameters} \\ 
\midrule
Gemma-2           \cite{gemma_2024}      & 9B                  \\
Llama-3         \cite{llama3modelcard} & 70B                    \\
Llama-3.3       \cite{meta2024llama}   & 70B                  \\
Llama-3       \cite{meta2024llama}     & 8B                \\
Mixtral       \cite{Jiang2024MixtralOE}     & 8x7B                  \\ 
\bottomrule
\end{tabular}
\caption{Models for experiments}
\label{tab:models}
\end{table}

We select models that are trained on publicly available online data. All LLMs are instructed in English, and we experiment with three different prompt templates, selecting the best-performing one for our experiments. To ensure consistency of the results, each experiment is conducted three times, with same temperature. We calculate variance across the three runs for both English and Telugu using the F1 scores of the best-performing models. For English, the Naive RAG exhibits the highest variance, while the Advanced RAG shows the lowest. For Telugu, Automatic Scraping results in the highest variance, whereas the Naive RAG has the lowest. Table \ref{tab:compact-model-justification} shows average scores of model performance across English and Telugu datasets. We use multiple evaluation metrics in our experiments to gain a comprehensive understanding of the models' performance, as no single metric can fully capture the quality of a model's output for justification generation.

\newcommand{\best}[1]{\underline{#1}}

\begin{table*}[htbp]
\centering
\footnotesize
\setlength{\tabcolsep}{3pt}
\renewcommand{\arraystretch}{1.1}
\resizebox{\textwidth}{!}{%
\begin{tabular}{@{}l|c|cc||ccccc|c||ccccc|c@{}}
\toprule
\multirow{3}{*}{Model} & Approach & \multicolumn{2}{c||}{F1 (Claim Verif.)} & \multicolumn{5}{c}{English Justification Generation Scores} &  & \multicolumn{5}{c}{Telugu Justification Generation Scores} & \\
\cmidrule(lr){3-4} \cmidrule(lr){5-10} \cmidrule(lr){10-10} \cmidrule(lr){11-16} \cmidrule(lr){16-16}
& & En & Te & METEOR & R-L & ChrF & BLEURT & BERTScore & Avg-En & METEOR & R-L & ChrF & BLEURT & BERTScore & Avg-Te \\
\midrule
\multirow{4}{*}{Llama-3-70B} 
& SP     & 80.16 & 42.95 & \best{0.288} & 0.283 & 39.92 & 0.48 & 0.87 & 0.464 & 0.126 & 0.165 & 25.34 & 0.45 & \best{0.72} & 0.343 \\
& N-RAG  & 58.16 & 40.77 & 0.267 & 0.275 & 38.51 & 0.47 & 0.86 & 0.451 & \best{0.140} & 0.163 & 23.94 & 0.44 & 0.71 & 0.339 \\
& A-RAG  & 61.21 & 44.31 & 0.256 & 0.259 & \best{\textbf{41.11}} & \best{\textbf{0.56}} & 0.88 & \best{0.473} & 0.134 & \best{0.174} & \best{26.08} & \best{0.48} & \best{0.72} & \best{0.354} \\
& AS     & \best{\textbf{86.14}} & \best{80.45} & 0.281 & \best{0.289} & 37.72 & 0.47 & \best{0.89} & 0.461 & 0.123 & 0.162 & 24.70 & 0.45 & \best{0.72} & 0.340 \\
\midrule
\multirow{4}{*}{Llama-3.3-70B} 
& SP     & 75.07 & 70.68 & 0.275 & 0.275 & 38.55 & 0.47 & 0.89 & 0.459 & \best{\textbf{0.172}} & \best{\textbf{0.229}} & \best{\textbf{32.79}} & \best{\textbf{0.51}} & 0.71 & \best{\textbf{0.390}} \\
& N-RAG  & 57.44 & 38.86 & 0.286 & 0.282 & 35.41 & 0.43 & 0.87 & 0.444 & 0.106 & 0.123 & 27.04 & 0.40 & 0.72 & 0.324 \\
& A-RAG  & 59.38 & 41.76 & 0.259 & 0.250 & 37.81 & 0.42 & 0.88 & 0.437 & 0.135 & 0.174 & 28.29 & 0.43 & 0.72 & 0.348 \\
& AS     & \best{77.22} & \best{\textbf{80.58}} &  \best{0.308} & \best{0.318} & \best{39.81} & \best{0.49} & \best{\textbf{0.90}} & \best{0.482} & 0.163 & 0.196 & 31.84 & 0.50 & \best{\textbf{0.73}} & 0.381 \\
\midrule
\multirow{4}{*}{Llama-3-8B} 
& SP     & 56.21 & 48.45 & \best{0.294} & 0.279 & \best{39.85} & \best{0.50} & \best{0.89} & \best{0.472} & 0.138 & 0.194 & \best{29.64} & 0.42 & \best{\textbf{0.73}} & 0.356 \\
& N-RAG  & 52.41 & 47.29 & 0.266 & 0.280 & 37.59 & 0.45 & 0.86 & 0.446 & \best{0.139} & 0.184 & 28.21 & 0.41 & 0.72 & 0.347 \\
& A-RAG  & 60.11 & 49.75 & 0.254 & \best{0.304} & 38.41 & 0.49 & \best{0.89} & 0.464 & 0.133 & \best{0.203} & 29.61 & \best{0.44} & 0.72 & \best{0.358} \\
& AS     & \best{70.83} & \best{50.77} & 0.288 & 0.291 & 38.64 & 0.47 & 0.87 & 0.461 & 0.124 & 0.164 & 25.96 & 0.42 & 0.72 & 0.338 \\
\midrule
\multirow{4}{*}{Mixtral-8x7B} 
& SP     & 56.95 & 49.22 & 0.285 & 0.273 & 38.94 & 0.49 & 0.88 & 0.463 & 0.110 & 0.129 & 27.59 & 0.29 & 0.72 & 0.305 \\
& N-RAG  & 57.19 & 51.24 & 0.293 & 0.303 & 37.51 & 0.47 & 0.86 & 0.460 & \best{0.153} & 0.172 & 28.39 & 0.41 & \best{\textbf{0.73}} & 0.350 \\
& A-RAG  & 59.26 & 55.49 & 0.280 & 0.293 & 38.66 & \best{0.51} & \best{0.89} & 0.472 & 0.146 & \best{0.213} & \best{28.66} & \best{0.43} & 0.72 & \best{0.359} \\
& AS     & \best{84.08} & \best{73.86} & \best{\textbf{0.316}} & \best{\textbf{0.340}} & \best{41.03} & 0.48 & 0.87 & \best{\textbf{0.483}} & 0.087 & 0.114 & 23.27 & 0.28 & 0.70 & 0.283 \\
\midrule
\multirow{4}{*}{Gemma-2-9B} 
& SP     & 64.72 & 57.41 & \best{0.208} & \best{0.283} & 31.89 & 0.46 & 0.87 & 0.428 & \best{0.125} & 0.183 & 26.66 & 0.43 & \best{\textbf{0.73}} & 0.347 \\
& N-RAG  & 62.21 & 52.39 & 0.197 & 0.264 & 30.51 & 0.45 & 0.87 & 0.417 & 0.103 & 0.173 & 28.41 & 0.43 & 0.72 & 0.342 \\
& A-RAG  & 63.81 & 50.77 & 0.180 & \best{0.283} & 34.74 & \best{0.49} & \best{\textbf{0.90}} & 0.440 & 0.094 & \best{0.213} & \best{30.49} & \best{0.46} & 0.72 & \best{0.358} \\
& AS     & \best{83.23} & \best{78.05} & 0.217 & 0.277 & \best{36.77} & 0.48 & 0.87 & \best{0.442} & 0.114 & 0.152 & 24.68 & 0.42 & 0.72 & 0.331 \\
\bottomrule
\end{tabular}
}
\caption{Scores across different metrics for English (En) and Telugu (Te). Approaches include Simple Prompting (SP), Naive RAG (N-RAG), Advanced RAG (A-RAG), and Automatic Scraping (AS). The best results for each metric and language are highlighted in \textbf{bold}, while the best scores per metric and language for each model are \underline{underlined}. ChrF scores are normalized (divided by 100) when computing average scores for English and Telugu.}
\label{tab:compact-model-justification}
\end{table*}

\begin{table}[h]
\centering
\small
\setlength{\tabcolsep}{6pt}
\begin{tabular}{lcc}
\toprule
\textbf{Model} & \textbf{En} & \textbf{Te} \\
\midrule
Llama-3-70B     & 0.101 & 0.072 \\
Llama-3.3-70B   & 0.140 & 0.090 \\
Llama-3-8B      & 0.178 & \textbf{0.124} \\
Mixtral-8x7B    & \textbf{0.208} & 0.089 \\
Gemma-2-9B      & 0.126 & 0.079 \\
\bottomrule
\end{tabular}
\caption{QA pairs Hungarian METEOR scores for English (En) and Telugu (Te). Best scores are highlighted in \textbf{bold}}
\label{tab:qa_pairs}
\end{table}

\section{Results and Discussion}

We analyze claim verification and justification generation results for English and Telugu to answer RQ1 and RQ2, and also analyze QA pair results.
\subsection{Claim Verification}


In order to answer RQ1, we examine the claim verification results presented in Table \ref{tab:compact-model-justification}.

\subsubsection{English}

\textbf{Simple Prompting:} Within the Simple Prompting approach across models, Llama-3-70B achieves the highest F1 score, likely due to its large size and English-focused training, enabling strong reasoning without external supporting documents. In contrast, Llama-3-8B performs the worst, likely due to its smaller size. Interestingly, Llama-3-70B outperforms both Naive and Advanced RAG under Simple Prompting, showing the largest performance gap of 23.95 points of F1 score between the best and worst performing models.

\textbf{Automatic Scraping:} We observe that all models obtain their highest F1 scores with this approach. With Automatic Scraping Llama-3-70B has the highest F1 score and Llama-3-8B has the lowest F1 score. This suggests that Automatic Scraping provides high-quality, relevant context that helps LLMs verifying and classifying the claims. All models perform better with Automatic Scraping compared to Simple Prompting.

\textbf{Naive RAG:} Gemma-2-9B achieves the highest F1 score and Llama-3-8B has the lowest F1 score. We observe that the Naive RAG approach does not improve the models' performance with respect to Simple Prompting except for Mixtral-8x7B. One possible reason for the relatively low F1 scores across models is that the Cohere model may not retrieve suitable supporting documents, particularly for claims related to the Indian context. This limitation at the evidence retrieval stage can significantly impact the quality of context available to the LLM, thus reducing overall performance. 


\textbf{Advanced RAG:} Gemma-2-9B achieves the highest F1 score and Mixtral-8x7B shows the lowest F1 score. We observe that models consistently perform slightly better with Advanced RAG compared to Naive RAG. This improvement may be attributed to the additional components in Advanced RAG that enhance the models' overall performance. However, results with the Simple Prompting approach remain superior except for Llama-3-8B and Mixtral-8x7B. Notably, this approach results in the smallest performance gap of 4.55 points in average F1 score between the best and worst performing models.

\subsubsection{Telugu}

\textbf{Simple Prompting:} Within the Simple Prompting approach, Llama-3.3-70B obtains the highest F1 score, likely due to some knowledge of Telugu in its pre-training data, as it was trained on open-source web documents. In contrast, all other models have low F1 scores. This could be due to the limited presence of Telugu in their pre-training corpora.

\textbf{Naive RAG:} Gemma-2-9B has the highest F1 score and Llama-3.3-70B has the lowest F1 score. The relatively low scores across models may be attributed to the Cohere model's limited ability to retrieve relevant supporting documents for claims in Telugu. Since Telugu is a low-resource language, the amount and quality of content available in it would be significantly lower compared to English. In this approach, only Mixtral-8x7B performs better than the models with Simple Prompting. This approach has the lowest performance gap of 14.03 F1 points between the best and the worst performing models.

 \textbf{Advanced RAG:}
 Mixtral-8x7B has the highest F1 score and Llama-3.3-70B has the lowest F1 score. The F1 scores across models suggest that the Advanced RAG generally performs slightly better than the Naive RAG for Telugu, with the exception of Gemma-2-9B. This exception may be due to Gemma-2-9B not having received suitable documents as context.
 The modest improvements seen with Advanced RAG can likely be attributed to its additional components. However, F1 scores for Telugu remain relatively low compared to those for English. Among the evaluated models, Llama-3-70B, Llama-3-8B, and Mixtral-8x7B outperform Simple Prompting.

\textbf{Automatic Scraping:} Under this method, in which the context is in English, Llama-3.3-70B achieves the highest F1 score, demonstrating its ability to transfer knowledge from English to Telugu. In comparison, the smaller Llama-3-8B has the lowest F1 score. These results highlight that LLMs perform significantly better in Telugu when provided with suitable supporting documents. Here, all models perform better than Simple Prompting. The performance gap between the best and the lowest performing model is 29.81 average F1 score, which is highest using this technique.

Automatic scraping has the highest scores for claim verification as it uses reliable supporting documents as context. To answer RQ1, our experiments show that LLMs perform better at claim verification in English compared to Telugu.

\subsection{Justification Generation}
As shown in Table \ref{tab:compact-model-justification}, we compare the results of justification generation score (JGS) for Telugu and English to answer RQ2. JGS is an average of METEOR, R-L, ChrF, BLUERT and BERTScore. We observe that for English and Telugu different models and approaches have high scores across different metrics. However, for English, Mixtral-8x7B with Automatic Scraping has the highest overall average JGS. The best overall JGS in Telugu is attained by Llama-3.3-70B using Simple Prompting. Manual review of 100 justifications from various methods reveals no clear link between claim verification and JGS.

\subsection{QA pairs}
As shown in Table \ref{tab:qa_pairs}, Mixtral-8x7B achieves the highest METEOR score for English, likely because it is trained on predominantly English data. In contrast, Llama-3-8B, despite being a small model, achieves the best METEOR score for Telugu. This performance may result from its closer adherence to the reference QA pairs, whereas larger models tend to ``hallucinate" or be creative~\cite{lin-etal-2022-truthfulqa}, which negatively affects similarity scores .

\section{Error Analysis} 
In this section, we present the qualitative and quantitative error analysis for English and Telugu.

\subsection{Qualitative Error Analysis}
We manually analyze 100 samples from the best performing models for each task: Llama-3-70B (English) and Llama-3.3-70B (Telugu) for claim verification; Mixtral-8x7B (English) and Llama-3.3-70B (Telugu) for justification generation.
\subsubsection{Claim Verification}

\begin{figure}[h!]
    \centering
    \fbox{\includegraphics[width=0.44\textwidth]{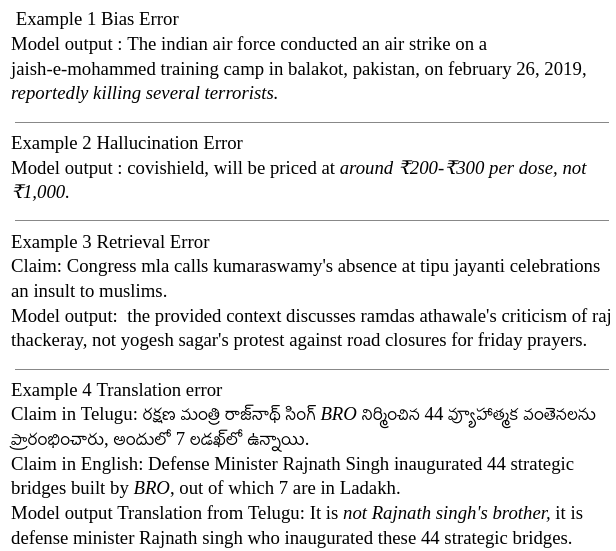}}
    \caption{Different types of Errors}
    \label{fig:Errors}
\end{figure}


We have focused on identification of biases \cite{dev-etal-2022-measures}, hallucinations \cite{li-etal-2024-dawn}, retrieval, and translation errors. \textbf{Biases} are unfair patterns in responses that occur when the model favors certain views, stereotypes, or groups over others. As shown in Example one in Figure \ref{fig:Errors}, there is a potential bias toward labeling individuals as terrorists. \textbf{Hallucinations} occur when LLMs generate information that is factually incorrect. In Example two in Figure \ref{fig:Errors} the language model hallucinates about the pricing of the Covishield vaccine. \textbf{Retrieval errors} in RAG approaches refer to the failing of the model to obtain relevant or sufficient contextual knowledge to support accurate reasoning, leading to incorrect or unsupported output. Example three in Figure \ref{fig:Errors} shows that the retrieved documents are not related to the claim about \textit{kumaraswamy} and \textit{tipu jayanti}. Finally, \textbf{translation errors} are uniquely observed when there is a language mismatch in the claim or between claim and context - for example, when there are acronyms in English and the claim is in Telugu. In such scenarios, the models attempt to translate the English acronyms to Telugu as in Example four in Figure \ref{fig:Errors} where it can be observed that BRO acronym which is in English is translated to ``brother'' in Telugu.

\subsection{Justification Generation}
We manually evaluate generated 100 justifications against the gold-standard justifications from different approaches. We observe that Automatic Scraping enables LLMs to generate good-quality justifications in English and Telugu. Manual inspection further reveals that the quality of text generation is generally good for English across different models and approaches. However, outputs in Telugu often exhibit syntactic and semantic errors, along with instances of Tenglish (a mix of Telugu and En-
glish) script.

\subsection{Quantitative Error Analysis}


We use the mistral-saba-24B LLM \cite{mistral2025small3} as a judge, following an in-context learning approach.We manually select one misclassified claim with its justification from each error type as a demonstration for the judge. Misclassified claims and their justifications are filtered and they are then classified by the LLM into the predefined error categories, with uncategorized errors labeled as ``Other." Tables \ref{tab:english-errors} and \ref{tab:telugu-errors} report category-wise error percentages. Manual verification of 50 errors per language confirms accurate quantification.



\begin{table}[t]
\centering
\scriptsize
\setlength{\tabcolsep}{8pt} 
\begin{tabular}{lcccc}
\toprule
\textbf{Approach} & \textbf{B} & \textbf{H} & \textbf{R} & \textbf{O} \\
\midrule
SP            & 13.14\%  & 4.81\%  & --          & 1.92\%  \\
AS           & 4.91\%   & 1.02\%   & 3.85\%      & 4.08\%  \\
N-RAG        & 12.12\%  & 5.39\%   & 13.54\%     & 10.76\% \\
A-RAG        & 11.98\%  & 1.22\%   & 16.75\%     & 9.27\% \\
\bottomrule
\end{tabular}
\caption{English errors with percentage (relative to 5006 claims). B: Biases, H: Hallucinations, R: Retrieval, O: Other.}
\label{tab:english-errors}
\end{table}

\begin{table}[t]
\centering
\scriptsize
\setlength{\tabcolsep}{6pt} 
\begin{tabular}{lccccc}
\toprule
\textbf{Approach} & \textbf{B} & \textbf{H} & \textbf{R} & \textbf{T} & \textbf{O} \\
\midrule
SP     & 5.17\% & 10.71\% & --  & -- & 13.16\% \\
AS       & 1.00\%  & 2.46\% & --  & 0.26\% & 12.86\% \\
N-RAG  & 8.79\% & 9.35\% & 1.62\%  & 8.63\% & 20.59\% \\
A-RAG   & 0.50\%  & 8.25\% & 10.53\% & --   & 13.18\% \\
\bottomrule
\end{tabular}
\caption{Telugu errors. B: Biases, H: Hallucinations, R: Retrieval, T: Translation, O: Other.}
\label{tab:telugu-errors}
\end{table}

\section{Conclusion}

In this project, we introduced a new English-Telugu claim verification dataset with manually annotated QA pairs and justifications. We used it to benchmark Simple Prompting and RAG approaches with LLMs. Our results show that the models perform better in English than in Telugu, highlighting challenges in claim verification and justification generation in Telugu.

\textbf{Limitations}

The results of our experiments are based on a dataset of 5,006 claims with only two labels from five topics. Performance may vary with larger and more diverse datasets. In India, claims occur in multiple languages, but for this study, we work in one language at a time. We need to explore different prompt templates for Telugu and English, as some templates perform better than others. Our dataset consists only of textual claims, excluding images and videos, which are also commonly associated with the spread of false claims. Although we have relied on lexical and semantic similarity metrics, we have not incorporated additional text generation metrics to detect hallucinations. Our evaluation relies exclusively on automatic metrics such as R-L, METEOR, and BERTScore. While these provide surface-level and semantic overlap, they may not adequately capture the true quality of either QA pairs or justifications. In particular, justifications can often be expressed in many valid ways that differ substantially from the reference, leading to artificially low metric scores, while conversely, outputs that are lexically or semantically similar to the reference may still be incorrect. The limited variance in our reported BERTScore values (0.70–0.73) for Telugu further suggests that these metrics may not be sensitive enough to meaningful differences in justification quality. A more robust assessment would require human evaluation, which could better judge correctness, faithfulness, and usefulness of both the questions/answers and the justifications. Future work should therefore complement automatic metrics with systematic human evaluation. Naive RAG and Advanced RAG approaches that we use for experiments often require significant processing time, particularly for languages like Telugu. This is due to the complexity of tokenization, retrieval, and generation stages, which may not be as optimized for low-resource languages as they are for English. We have used RSS feeds from only a small number of sources and we have not performed ablation studies on the individual components of Advanced RAG. Since our dataset is derived through translation from English, it may not fully represent native Telugu. Translations tend to exhibit different levels of formality, topic distribution, and cultural biases compared to texts in Telugu produced by native speakers. Therefore, while our dataset serves as a useful resource, we acknowledge that future work should prioritize collecting and incorporating more native-authored Telugu data.

\section*{Acknowledgements}

We thank Begari Kaveri, Sujatha Theetla and Ravi Teja Chikkala for reviewing and editing the Telugu translations and inter-annotation agreement of the Preethi dataset.

This project was supported by the German Federal Ministry of Research, Technology and Space (BMFTR) as part of the project TRAILS (01IW24005), by \textit{DisAI - Improving scientific excellence and creativity in combating disinformation with artificial intelligence and language technologies}, a project funded by Horizon Europe under \href{https://doi.org/10.3030/101079164}{GA No.101079164}, by the \textit{European Union NextGenerationEU} through the Recovery and Resilience Plan for Slovakia under the project No. 09I01-03-V04-00007, and by Saarland University and University of Basque country in collaboration with Erasmus Mundus Language and Communication Technologies (EMLCT).

\bibliographystyle{acl_natbib}
\bibliography{anthology,ranlp2025}

\clearpage
\onecolumn
\appendix

\end{document}